\documentclass[10pt, conference]{IEEEtran}
\usepackage[para,online,flushleft]{threeparttable}
\usepackage{graphicx}
\usepackage{balance}  
\usepackage{url}
\usepackage{booktabs}
\usepackage{etoolbox}
\usepackage{tabularx}
\makeatletter
\patchcmd{\maketitle}{\@copyrightspace}{}{}{}
\makeatother

\let\labelindent\relax
\usepackage{amsmath}
\usepackage{amssymb}
\usepackage{leftidx}
\usepackage{bm}
\usepackage{comment}
\usepackage{color}
\usepackage{cite}
\usepackage{enumitem}
\usepackage{amsthm}
\usepackage[numbers]{natbib}
\usepackage[ruled, vlined, linesnumbered]{algorithm2e}
\usepackage{multicol}
\usepackage{graphicx}
\usepackage{capt-of}
\usepackage{etoolbox}
\makeatletter
\patchcmd{\@makecaption}
  {\scshape}
  {}
  {}
  {}
\makeatother

\usepackage{subfigure}
\usepackage{multirow}
\usepackage{cite}
\usepackage{colortbl}

\theoremstyle{definition}

\newtheorem{example}{Example}[section]
\theoremstyle{remark}
\theoremstyle{problem}

\newtheoremstyle{problemstyle}  
        {3pt}                                               
        {3pt}                                               
        {\normalfont\itshape}                               
        {}                                                  
        {\bfseries}                 
        {\normalfont\bfseries:}         
        {.5em}                                          
        {}                                                  
\theoremstyle{problemstyle}
\newtheorem{problem}{Problem}

\makeatletter
\newcommand*{\transpose}{%
  {\mathpalette\@transpose{}}%
}
\newcommand*{\@transpose}[2]{%
  \raisebox{\depth}{$\m@th#1\intercal$}%
}
\makeatother

\begin{document}
\title{Deep Learning for Predicting Dynamic Uncertain Opinions in Network Data}
\author{\IEEEauthorblockN{Xujiang Zhao, Feng Chen}
\IEEEauthorblockA{Computer Science Department\\
University at Albany -- SUNY, Albany, NY, USA\\
xzhao8@albany.edu, fchen5@albany.edu}
\and
\IEEEauthorblockN{Jin-Hee Cho}
\IEEEauthorblockA{Department of Computer Science\\
Virginia Tech, Falls Church, VA, USA\\
jicho@vt.edu}\thanks{This work is done when Jin-Hee Cho was with US Army Research Laboratory.}
}
\maketitle

\begin{abstract} 

Subjective Logic (SL) is one of well-known belief models that can explicitly deal with uncertain opinions and infer unknown opinions based on a rich set of operators of fusing multiple opinions. Due to high simplicity and applicability, SL has been substantially applied in a variety of decision making in the area of cybersecurity, opinion models, trust models, and/or social network analysis. However, SL and its variants have exposed limitations in predicting uncertain opinions in real-world dynamic network data mainly in three-fold: (1) a lack of scalability to deal with a large-scale network; (2) limited capability to handle heterogeneous topological and temporal dependencies among node-level opinions; and (3) a high sensitivity with conflicting evidence that may generate counterintuitive opinions derived from the evidence. In this work, we proposed a novel deep learning (DL)-based dynamic opinion inference model while node-level opinions are still formalized based on SL meaning that an opinion has a dimension of uncertainty in addition to belief and disbelief in a binomial opinion (i.e., agree or disagree). The proposed DL-based dynamic opinion inference model overcomes the above three limitations by integrating the following techniques: (1) state-of-the-art DL techniques, such as the Graph Convolutional Network (GCN) and the Gated Recurrent Units (GRU) for modeling the topological and temporal heterogeneous dependency information of a given dynamic network; (2) modeling conflicting opinions based on robust statistics;  and (3) a highly scalable inference algorithm to predict dynamic, uncertain opinions in a linear computation time. We validated the outperformance of our proposed DL-based algorithm (i.e., GCN-GRU-opinion model) via extensive comparative performance analysis based on four real-world datasets. 
\end{abstract}

\section{Introduction}
In the decision making domain, including the fields of evidence and belief theories, reasoning or managing uncertainty has been studied since 1960s. The examples include Fuzzy Logic, Dempster-Shafer Theory (DST), Transferable Belief Model, and Dezert-Smarandache Theory~\cite{chen2017collective}. These theories deal with uncertainty implicitly. In 1990's, as another variant of DST, Subjective Logic (SL)~\cite{josang2001logic} is proposed to deal with a dimension of uncertainty in subjective opinions more explicitely. SL defines a binomial opinion (e.g., agree vs. disagree) with three dimensions, including belief, disbelief, and uncertainty. SL provides a set of various operators to fuse multiple, different opinions that allow deriving structural relations between opinions in a network. Although SL has offered a rich set of fusion operators, its inherent parametric way of combining opinions has been shown as a hurdle to limit its scalability and led to a bounded prediction accuracy in deriving unknown opinions. To handle these issues, the variants of SL have been proposed to resolve the issue of scalability in SL, such as subjective Bayesian networks~\cite{ivanovska2015subjective} or collective subjective logic based on Markov Random Fields (MRFs)~\cite{chen2017collective}. However, due to the inherent parametric opinion derivation using fusion operators and the distribution assumption (e.g., Bayesian networks) in SL, the bounded performance of SL and its variants~\cite{ivanovska2015subjective, chen2017collective} have not been resolved.

This paper focuses on predicting  binomial opinions in a real-world dynamic network data. In particular, each node in a dynamic network has a random opinion at a different time stamp that is either observed or unknown. Assuming that the opinions of some nodes are observed at time $t$ snapshot, we aim to predict the unknown node-level binomial opinions among all the snapshots. To this end, we should resolve the following challenges: 
\begin{enumerate}
\item \textbf{Scalability:} The traditional SL method scales exponentially as the number of nodes and/or edges increases because it should identify shortest paths to fuse known opinions to derive unknown opinions. In this sense, we aim to develop a linear complexity algorithm that scales well as the total number of nodes and/or edges increases.  
\item \textbf{Heterogeneity:} There often exist heterogeneous dependencies among node-level opinions across topological and time spaces of real-world networks. However, the original SL model and its variants treat heterogeneous data and network structure homogeneously with the assumption of parametric statistical models, such as Bayesian networks or MRFs, which may not resolve the issue introduced by data and dependency heterogeneity 
\item \textbf{Sensitivity to opinions derived based on conflicting evidence:} Some variants of SL have been proposed to deal with conflicting evidence in order not to reduce uncertainty in the presence of conflicting evidence. For example, recently an SL-based opinion update model is proposed to readjust its derived opinion when conflicting opinions are received~\cite{Cho17}. However, how to deal with conflicting evidence in deriving unknown opinions with known dynamic, uncertain opinions has not been studied in the existing approaches.  
\end{enumerate}

In this work, we developed a novel deep learning (DL)-based model to predict unknown opinions in dynamic network data. In particular, a node-level opinion is formulated as SL-based binomial opinion, consisting of belief, disbelief, and uncertainty masses. Our proposed DL-based dynamic opinion inference model captures the topological and temporal information of node-level beliefs and uncertainties explicitly and effectively and efficiently addresses the above three challenges by considering (1) state-of-the-art DL techniques, such as the \textbf{Graph Convolutional Network} (GCN) and the \textbf{Gated Recurrent Units} (GRU) for modeling the topological and temporal heterogeneous dependency information of a given dynamic network; (2) \textbf{robust modeling of conflicting opinions} based on robust statistics that eliminate the bias introduced by conflicting opinions, treating them as outliers; and (3) \textbf{a  highly scalable inference algorithm} for predicting dynamic uncertain opinions that scale in a linear complexity, proportional to the size of a network (i.e., the number of nodes). This work makes the following \textbf{key contributions}:
\begin{enumerate}
    \item This work is the first that explores DL-techniques, such as GCN and GRU, for modeling the heterogeneous dependency information of uncertain opinions in real-world dynamic networks. We conducted an extensive experimentation in order to validate the efficiency and effectiveness of the proposed DL models based on four real-world network datasets. 
    \item Our proposed DL-based dynamic opinion inference model well integrates GCN, GRU, and a robust statistical layer. This is the first that can address the three key challenges of dynamic uncertain opinion prediction simultaneously, associated with scalability, heterogenity, and sensitivity to conflicting evidence. 
    \item We validated the proposed DL-based dynamic opinion inference algorithm through extensive experiments using three real-world and one semi-synthetic benchmark datasets. We compared the performance of our proposed DL-based approach with those of the original SL and the state-of-the-art counterpart (i.e., collective subjective logic, or CSL~\cite{chen2017collective}), as well as traditional DL methods, including GCN and GRU. The implementation of our proposed methods and the tested datasets will be available at github after the paper is accepted. 
\end{enumerate}

\section{Related Work} \label{sec: related_work}

\subsection{Probabilistic Models} \label{subsec:probabilistic-model-related}
Uncertainty reasoning and modeling caused by \textit{a lack of information} or {\em knowledge} in network data has been substantially studied as a joint probability distribution over a set of variables, in which each variable relates a node in a given network. Two typical probabilistic models include MRFs~\cite{pearl2014probabilistic} and Gaussian Processes (GPs)~\cite{rasmussen2004gaussian}. The former models the joint distribution based on potential functions of the cliques to capture the relational structure. The latter models the joint distribution using a multivariate Gaussian distribution and uses the covariance matrix to characterize the structural relations between the variables in the network. 

Probabilistic models have shown limited capabilities in considering uncertainty caused by ignorance (i.e., a lack of evidence about the ground truth states) and other causes, such as vagueness (i.e., failing in discerning a single state) and ambiguity (i.e., failing in observing consensus due to conflicting evidence). For example, if somebody wants to express ignorance about the state $x$ as ``\textit{I don't know},'' this would be impossible with a simple probability value. A probability $P(x) = 0.5$ would mean that $x$ and $\bar{x}$ (i.e., $1-x$) are equally likely, which is quite informative in deed, unlike ignorance. 

\subsection{SL-based Opinion Inference Models} \label{subsec:sl-based models}

SL has been proposed to define an opinion that explicitly deals with uncertainty. In addition, SL offers a variety of operators to fuse multiple opinions~\cite{josang2016subjective}. New extensions of SL have been proposed to make SL scalable to large-scale networks, such as subjective Bayesian networks~\cite{ivanovska2015subjective} and collective subjective logic (CSL), as a hybrid approach, by combining SL, probabilistic soft logic, and MRFs~\cite{chen2017collective, bach2015hinge}. However, all the preceding belief models are designed based on predefined operators or distribution assumptions (e.g., Bayesian networks) that may not effectively deal with heterogeneous uncertain opinions in given dynamic network data. In this work, we proposed a DL-based opinion inference model based on GCN and GRU which maximizes prediction accuracy while minimizing computation time. This way allows us to effectively and efficiently infer unknown opinions under large network data where we still retain the merit of SL's uncertain opinions in dealing with heterogeneous node-level opinions.
 
\subsection{DL-based Inference Models} \label{subsec:dl-inference-related}
In early days of machine learning (or deep learning), recurrent neural networks (RNNs) are used to deal with data representations in directed acyclic graphs~\cite{frasconi1998general}. Later, Graph Neural Networks (GNNs)~\cite{gori2005new} are developed as a generalization of RNNs to process general directed and undirected graphs. After then, convoluational neural networks (CNNs) is developed to deal with data representations from a spatial domain to a graph domain. The methods developed in this direction are called graph convoluational networks (GCNs) and fall into two main categories: spectral approaches and non-spectral approaches. GCNs have demonstrated the state-of-the-art performance in a number of challenging mining tasks (e.g., semi-supervised node classification and link prediction)~\cite{DBLP:journals/corr/KipfW16, hamilton2017inductive}. 

{\em Spectral approaches} for GCNs explore convolutions based on a spectral representation of the graphs. \citet{bruna2013spectral} implemented the convolution operator as a spectral filter in the Fourier domain by calculating the eigen-decomposition of the graph Laplacian, which, however, is computationally expensive and leads to non-spatially localized filters. \citet{henaff2015deep} proposed a parameterization of the spectral filters to make them spatially localized. 

\section{Background}  \label{sec:background}
For this work to be self-contained, this section provides the overview of SL, GCN, and GRU that are used to propose the DL-based opinion inference model in this work.

\subsection{Subjective Logic (SL)} \label{subsec:sl}

In SL, a binomial opinion is defined in terms of belief, disbelief, and uncertainty towards a given proposition $x$. For simplicity, we omit $x$ in the following notations~\cite{josang2016subjective}. To formally put, an opinion $w$ is represented by $w  =  (b, d, u, a)$ where $b$ is belief (e.g., true), $d$ is disbelief (e.g., false), and $u$ is uncertainty (i.e., ignorance or lack of evidence). $a$ represents a base rate, a prior knowledge upon no commitment (i.e., neither true nor false), where $b + d + u = 1$ for $(b, d, u, a) \in [0, 1]^4$. We denote an opinion by $w$, which can be \textit{projected} onto a single probability distribution by removing the uncertainty mass. 

A binomial opinion follows a Beta pdf (probability density function), denoted by $\text{Beta}(p | \alpha, \beta)$, where $\alpha$ represents the amount of positive evidence and $\beta$ is the amount of negative evidence~\cite{josang2016subjective}. In SL, uncertainty $u$ decreases as more evidence, $\alpha$ and $\beta$, is received over time. An opinion $w$ can be obtained based on $\alpha$ and $\beta$ as $w = (\alpha, \beta)$. This can be translated to $w = (b, d, u, a)$ using the mapping rule in SL.

SL offers an operator, $\otimes$, to discount trust when an entity does not have any direct experience with another entity. That is, transitive trust based on structural relations is used to derive trust between two entities who have not interacted before. Trust from $i$ to $j$, denoted by $w^i_j = (b^i_j, d^i_j, u^i_j, a^i_j)$, and trust from $j$ to $k$, $w^j_k = (b^j_k, d^j_k, u^j_k, a^j_k)$, are used to derive trust from $i$ to $k$, $w^i_k:= (b^i_k, d^i_k, u^i_k, a^i_k) = w^i_j \otimes w^j_k$. 
The well-known discounting operator, $\otimes$, to weigh trust of another entity's opinion and consensus operator, $\oplus$, to fuse two different opinions are used to derive trust measures based on the trust opinions of relationships~\cite{josang2016subjective}. Due to space constraint, we omitted the details of those operators and an example scenario using the operators. Interested readers can be referred to~\cite{josang2016subjective, chen2017collective}.

In this work, we aim to derive a set of unknown opinions $\mathbf{x} = \{x_1, \cdots, x_n\}$ when a set of observed opinions $\mathbf{y} = \{y_1, \cdots, y_m\}$ is given where both opinions are represented by a binomial opinion with the four dimensions described earlier (i.e., $w  =  (b, d, u, a)$), for example, $w_{x_i}$ for $i= 1 \cdots n$ and $w_{y_j}$ for $j= 1 \cdots m$. 

\subsection{Graph Convolutional Networks (GCN)} \label{subsec:gcn}
A GCN model~\cite{DBLP:journals/corr/KipfW16} works as follows. Denote a graph as $\mathbb{G} = (\mathbb{V}, \mathbb{E}, \textbf{A})$, where $\mathbb{V} = \{1, \cdots, n\}$ refers to the set of nodes and $\mathbb{E} \subseteq \mathbb{V}\times \mathbb{V}$ refers to the set of edges. Let $\textbf{A}\in \{0, 1\}^{n\times n}$ be the adjacency matrix, where $A_{i, j} = 1$ if $(i,j) \in \mathbb{E}$ and, otherwise, $A_{i, j} = 0$. The (unnormalized) graph Laplacian matrix is an $n\times n$ symmetric positive-semidefinite matrix $\textbf{L} = \textbf{D} - \textbf{A}$, where $\textbf{D}$ is the degree matrix and $D_{i,i}$ refers to the degree of node $i$ and $D_{i,i} = 0$ for $i\neq j$. 

The graph Laplacian has an eigen decomposition $\textbf{L} = \Phi \Lambda \Phi^T$, where $\Phi = (\bm{\phi}_1, \cdots, \bm{\phi}_n)$ are the orthonormal eigenvectors and $\Lambda = diag(\lambda_1, \cdots, \lambda_n)$ is the diagonal matrix of corresponding eigenvalues. The eigenvalues serve as the role of frequencies  in classical harmonic analysis and the eigenvectors are interpreted as Fourier atoms. Given a signal ${\bf r}\in \mathbb{R}^n$ (or a vector of feature values) on the nodes of graph $\mathbb{G}$, where $r_i$ refers to a feature value at node $i$, its graph Fourier transform is given by $\hat{\bf r} = \Phi^T {\bf r}$. Given two signals ${\bf r}$ and ${\bf b}$ on the graph, we can define their spectral convolution as the element-wise product of their Fourier transformations, 
\begin{eqnarray}
{\bf r}\star {\bf b} = \Phi^T (\Phi^T {\bf r})\circ (\Phi^T {\bf b}) = \Phi diag(\hat{r}_1, \cdots, \hat{r}_n) \hat{\bf b},\label{spectral-convolution}
\end{eqnarray}
which is a property of the well-known {\em Convolutional Theorem} in the Euclidean case. 

As a graph is irregular  with nodes having widely different degrees, it is difficult to directly define a convolution on the nodes. Instead, \citet{bruna2013spectral} used the spectral definition of convolution (see Eq.~\eqref{spectral-convolution}) to generalize Convolutional Neural Networks (CNNs) on graphs, which has a spectral convolutional layer of the form as:
\begin{eqnarray}
g_\theta \star {\bf r} = \Phi g_\theta \Phi^T {\bf r}. \label{graph-convolution}
\end{eqnarray}
The filter $g_\theta$ can be defined as a function of the eigenvalues of $L$, i.e., $g(\Lambda)$. Evaluating Eq.~\eqref{graph-convolution} is computationally expensive because multiplication with the eigenvector matrix $\Phi$ is $O(n^2)$, in addition to the high computational cost in computing the eigendecomposition of $L$ in the first place. To address this problem, \citet{Hammond11} found that $g_\theta(\Lambda)$ can be well-approximated by a truncated expansion in terms of Chebyshev polynomials $T_k(r)$ up to $L$-th order: 
\begin{eqnarray}
g_\theta (\Lambda) \approx \sum\limits_{k=1}^L \theta_k T_k(\tilde{\Lambda}),
\end{eqnarray}
with a rescaled $\tilde{\Lambda} = \frac{2}{\lambda_{max}} \Lambda - I_n$. $\lambda_{max}$ refers to the largest eigenvalue of $L$. $\theta \in \mathbb{R}^L$ is a vector of Chebyshev coefficients. The Chebyshev polynomial can be recursively defined as $T_k(r) = 2x T_{k-1}(r) - T_{k-2}(r)$, with $T_0(r) = 1$ and $T_1(r) = r$. Applying the approximation based on Chebyshev polynomials, a convolution of a signal $x$ with a filter $g_\theta$ now has the approximated form: 
\begin{eqnarray}
g_\theta \star {\bf r} \approx \sum_{k=1}^K \theta_k T_k(\tilde{L}){\bf r}. \label{graph-convolution1}
\end{eqnarray}
By stacking multiple convolutional layers of the form of Eq.~\eqref{graph-convolution1} in which each layer is followed by a point-wise non-linearity filter, we can therefore design a multi-layer convolutional neural network model based on graph convolutions. 

\subsection{Gated Recurrent Units (GRU)} \label{subsec:gru}
The Gated Recurrent Units (GRU)~\cite{DBLP:journals/corr/ChoMGBSB14} is a variant of RNNs that can capture dependencies of different time stamps. The GRU uses trained gate units on inputs or memory states to keep the memory for a longer period of time which enables them to capture longer term dependencies than RNNs. Although the long short-term memory (LSTM) unit also uses gating  units and memory cell to model long-term dependencies, GRU is more efficient and has a less complex structure than LSTM. 

More formally, GRU is used to model the joint input sequence $\textbf{r} = (\textbf{r}^{(1)}, \cdots, \textbf{r}^{(T)})$ and output sequence $\textbf{h} = (\textbf{h}^{(1)}, \cdots, \textbf{h}^{(T)})$. 
GRU unit starts with calculating the \emph{update gate} $\textbf{z}^{(t)}$ at time stamp $t$, which decides how much of the past information (from previous time steps) needs to be passed along to the future: 
\begin{eqnarray}
\textbf{z}^{(t)} = \sigma(\textbf{W}_\textbf{z} \textbf{r}^{(t)} + \textbf{U}_\textbf{z} \textbf{h}^{t-1}),
\end{eqnarray}
where $\textbf{h}^{t-1}$ is the previous memory state, $\textbf{W}_\textbf{z}$ and $\textbf{U}_\textbf{z}$ is the weight parameters, $\textbf{r}^{(t)}$ is current feature input, and $\sigma$ is a sigmoid activation function.
The \emph{reset gate} $\textbf{s}^{(t)}$ is used from the model to decide how much of the past information to forget, the reset gate is computed by:
\begin{eqnarray}
\textbf{s}^{(t)} = \sigma(\textbf{W}_\textbf{s} \textbf{r}^{(t)} + \textbf{U}_\textbf{s} \textbf{h}^{t-1}). 
\end{eqnarray}
The current memory state $\tilde{\textbf{h}}^{(t)}$ will use the reset gate to store the relevant information from the past. It is calculated as follow:
\begin{eqnarray}
\tilde{\textbf{h}}^{(t)} = \tanh (\textbf{W} \textbf{r}^{(t)} + \textbf{U}(\textbf{s}^{(t)}\odot \textbf{h}^{(t-1)}))
\end{eqnarray}
where $\odot$ is an element-wise multiplication. The final memory state $\textbf{h}^{(t)}$ at time $t$ is a linear interpolation between the previous memory state $\textbf{h}^{(t-1)}$ and current memory state $\tilde{\textbf{h}}^{(t)}$:
\begin{eqnarray}
\textbf{h}^{(t)} =  \textbf{z}^{(t)} \odot {\textbf{h}}^{(t-1)} + (1-\textbf{z}^{(t)}) \odot \tilde{\textbf{h}}^{(t)} 
\end{eqnarray}
It determines what to collect from the current memory state $\tilde{\textbf{h}}^{(t)}$ and the previous memory state $\textbf{h}^{(t-1)}$. 

\begin{figure}[!htb]
  \begin{center}
  \vspace{-2mm}    \includegraphics[width=0.48\textwidth]{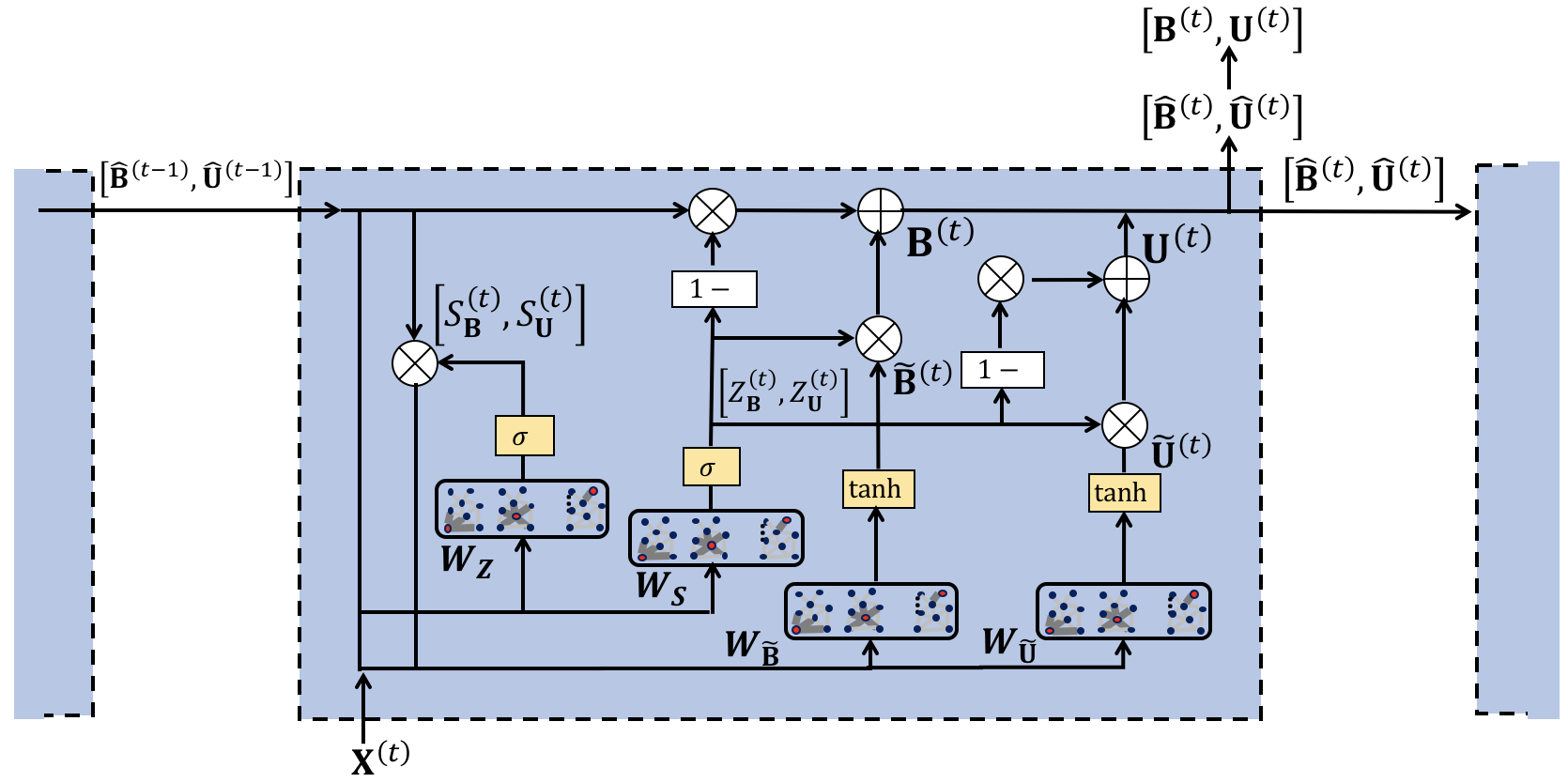}
  \end{center}
  \vspace{-4.5mm}
  \caption{The overview of our proposed DL-based dynamic opinion model.}
  \vspace{-3mm}
\label{fig:vae-opinion-model}
\end{figure}

\section{Problem Formulation} \label{sec:problem-formulation}
In this section, we describe an example to motivate a problem to solve in this work. We also show how to formulate a given uncertainty-based opinion inference problem.
\subsection{Example Scenario} \label{sect:exp}
In this work, we aim to infer unknown opinions, given a set of known opinions, in terms of the applications in traffic congestion prediction in a road network. Given a network, defined as $\mathbb{G} = (\mathbb{V}, \mathbb{E}, y)$, where $\mathbb{V} = \{1, 2, \cdots, N\}$ is the set of vertices (i.e., intersections in the road network), $\mathbb{E}\subseteq \mathbb{V}\times \mathbb{V}$ is the set of edges (i.e., road links), and  $y_i$ refers to a Boolean variable at node $i \in \mathbb{V}$, in which state $0$ indicates `non-congested' while state $1$ refers to `congested.' 

Suppose that at each time $t \in \{1, \cdots, T\}$,  we are given the subjective opinions of the congestion variables $\{y_i\}_{i \in \mathbb{L}_t}$, $\bm{\omega}_{\mathbb{L}_t}^{(t)} = [\omega_i^{(t)}]_{i \in \mathbb{L}_t}$ that are estimated based on their historical observations, where $\mathbb{L}_t \subseteq \mathbb{V}$ refers to a subset of nodes and $T$ is the total number of time stamps. 
A subjective opinion $\omega_i$ is defined by a tuple of three components, $\omega_{i} = (b_i, d_i, u_i)$. We assume that the subset of nodes $\mathbb{L}_t$ that has known opinions varies over time.

Given the information, we aim to predict the beliefs about the states of the congestion variables at the nodes without sensors (i.e., intersections without any camera) at each time $t \in \{1, \cdots, T\}$, denoted by $\{\bm{\omega}^{(t)}_{\mathbb{V}\setminus \mathbb{L}_{t=1}}, \cdots, \bm{\omega}^{(t)}_{\mathbb{V}\setminus \mathbb{L}_{t=T}}\}$, where $\bm{\omega}^{(t)}_{\mathbb{V}\setminus \mathbb{L}_t} =[\omega_i^{(t)}]_{i \in \mathbb{V} \setminus \mathbb{L}_t}$.  

\subsection{Problem Statement} \label{subsec:problem-statement}
The problem of uncertainty-based inference we aim to solve is formulated by: 
\begin{problem}[{\bf Uncertainty-based opinion inference in network data}] Let us define the following notations:
\begin{itemize}
\item Let $\mathbb{G} = (\mathbb{V}, \mathbb{E}, y)$ be an input network as defined above.  
\item Let $\omega_i^{(t)} = (b_i^{(t)}, d_i^{(t)}, u_i^{(t)})$ be node $i$'s subjective opinion of variable $y_i$ at time $t$, associated with node $i \in \mathbb{V}$ . Let $\mathbb{L}_t \subseteq \mathbb{V}$ be a subset of nodes whose opinions at time $t$ are denoted by $\bm{\omega}_\mathbb{L}^{(t)} = [\omega_{i}^{(t)}]_{i \in \mathbb{L}_t}$. 
\end{itemize}  
\noindent {\bf Given}
\begin{itemize}
\item $\mathbb{G} = (\mathbb{V}, \mathbb{E}, y)$, an input network; 
\item $\{\bm{\omega}^{(t=1)}_{\mathbb{L}_{t=T}}, \cdots, \bm{\omega}^{(t=T)}_{\mathbb{L}_{t=T}}\}$, a set of vectors of observed subjective opinions for $T$ times. 
\end{itemize}
{\bf Predict} $\{\bm{\omega}^{(t=1)}_{\mathbb{V}\setminus \mathbb{L}_{t=1}}, \cdots, \bm{\omega}^{(t=T)}_{\mathbb{V}\setminus \mathbb{L}_{t=T}}\}$, unknown opinions for $T$ time stamps. 
\end{problem}

\section{DL-Based Dynamic Opinion Model}
\label{sec:dl-opinion-inference-model}

In this section, we discuss the proposed DL-based dynamic opinion model and how it can be used to predict unknown dynamic opinions. Denote by $\textbf{B}^{(t)} = [b_i^{(t)}]_{i \in \mathbb{V}}$ and $\textbf{U}^{(t)} = [u_i^{(t)}]_{i \in \mathbb{V}}$ the vectors of the belief and uncertainty masses on the nodes in $\mathbb{V}$ at time $t$, respectively. The observed belief and uncertainty masses at time $t$ are denoted as $\textbf{B}^{(t)}_{\mathbb{L}_t} = [b_i^{(t)}]_{i \in \mathbb{L}_t}$ and $\textbf{U}^{(t)}_{\mathbb{L}_t} = [u_i^{(t)}]_{i \in \mathbb{L}_t}$, respectively. The unknown belief and uncertainty masses at time $t$ are denoted as $\textbf{B}^{(t)}_{\mathcal{V} \setminus \mathbb{L}_t}$ and $\textbf{U}^{(t)}_{\mathcal{V} \setminus \mathbb{L}_t}$, respectively. 


\subsection{Robust Statistical Opinion Derivation with Conflicting Opinions}
We consider a dynamic and potentially hostile scenario that some of the nodes may be compromised and their observed opinions may not reflect their true opinions and often conflict with the opinions of their neighboring nodes. Our key idea is to design an additional layer based a robust distribution of noises, i.e., Laplacian distribution, to alleviate the impact of conflicting opinions by treating them as outliers. In particular, the observed beliefs and uncertainties  ($\textbf{B}^{(t)}_{\mathbb{L}_t}$ and $\textbf{U}^{(t)}_{\mathbb{L}_t}$) are considered as a noise version of their true values ($\hat{\textbf{B}}^{(t)}_{\mathbb{L}_t}$ and $\hat{\textbf{U}}^{(t)}_{\mathbb{L}_t}$): 
\begin{eqnarray}
b^{(t)}_i &=& \hat{b}^{(t)}_i + e^{(t)}_i,\ e^{(t)}_i \sim \text{Lap}(\tau_b) \label{m1}\\
u^{(t)}_i &=& \hat{b}^{(t)}_i + r^{(t)}_i,\ r^{(t)}_i \sim \text{Lap}(\tau_u)\label{m2}
\end{eqnarray}
where $\text{Lap}(\nu_b)$ refers to a Laplacian distribution and its probability density function (PDF) is defined as $\text{Prob}(x) = \frac{1}{\nu_b} \exp{(-\frac{|x|}{\nu_b})}$, and $\tau_b$ and $\tau_u$ are the parameters of the two belief and uncertainty-related Laplacian distributions, respectively.  In this design, the error terms $e_i^{(t)}$ and $r_i^{(t)}$ will be non-zero if their related beliefs and uncertainties are from compromised users (or outliers) and conflicting with the related values of its neighbors; otherwise, they will be weighted as zero. 

\subsection{A GCN-GRU-based Module for Modeling High-Order Heterogeneous Dependencies among Node-Level Dynamic Uncertain Opinions} 

In the following, we combine GCN and GRU to jointly model the graph-structural and temporal dependencies among node-level beliefs $\{\hat{\bf B}^{(t)}\}_{t=1}^T$ and uncertainties $\{\hat{\bf U}^{(t)}\}_{t=1}^T$.  First, following the underlying principles of GRU, the beliefs $\hat{\bf B}^{(t)}$ and uncertainties $\hat{\bf U}^{(t)}$ at time $t$ are modeled as linear interpolations between the previous beliefs $\hat{\bf B}^{(t-1)}$ and uncertainties $\hat{\bf U}^{(t-1)}$ and the candidate values $\tilde{\bf B}^{(t)}$ and $\tilde{\bf U}^{(t)}$, respectively: 
\begin{eqnarray}
\hat{\bf B}^{(t)} =  {\bf Z}^{(t)}_{\bf B} \odot \hat{\bf B}^{(t-1)} + (1-{\bf Z}_{\bf B}^{(t)}) \odot \tilde{{\bf B}}^{(t)},\label{m3}\\
\hat{\bf U}^{(t)} =  {\bf Z}^{(t)}_{\bf U} \odot \hat{\bf U}^{(t-1)} + (1-{\bf Z}^{(t)}_{\bf U}) \odot \tilde{{\bf U}}^{(t)},\label{m4}
\end{eqnarray}
where \textbf{the update gates} ${\bf Z}^{(t)}_{\bf B}$ and ${\bf Z}^{(t)}_{\bf U}$ are trained on the current memory states $(\hat{\bf B}^{(t-1)}, \hat{\bf U}^{(t-1)})$ to keep the memory for a longer period of time thus enabling them to \textbf{capture longer term dependencies for both node-level beliefs and uncertainties}. 

The \textbf{update gates} ${\bf Z}^{(t)}_{\bf B}$ and ${\bf Z}^{(t)}_{\bf U}$ are computed by 
\begin{small}
\begin{eqnarray}
\vspace{-5mm}
\left[{\bf Z}_{\bf B}^{(t)}, {\bf Z}_{\bf U}^{(t)}\right] = \sigma\Big(\Big[g_{\textbf{W}_\textbf{Z}} \star [\hat{\bf B}^{(t-1)}, {\bf X}^{(t)}],g_{\textbf{W}_\textbf{Z}} \star [\hat{\bf U}^{(t-1)}, {\bf X}^{(t)}]\Big]\Big),
\label{update gates}
\vspace{-5mm}
\end{eqnarray}
\end{small} 
where $g_{\textbf{W}_\textbf{Z}} \star$ is a graph convolution defined in Eq.~\eqref{graph-convolution1},    $\textbf{W}_\textbf{Z}$ refers to the corresponding parameters, and \textbf{the matrix $\textbf{X}^{(t)}\in \mathbb{R}^{n\times 2}$ represents the current state of beliefs and uncertainties at time $t$, in which $X_{i,0}^{(t)} = b_i^{(t)}$ and $X_{i,1}^{(t)} = u_i^{(t)}$, if $i \in \mathbb{L}_t$; otherwise, $X_{i,0}^{(t)} = X_{i,1}^{(t)}= 0$  meaning that the belief and uncertainty are missing for node $i$ at time $t$}. 
This procedure takes a linear sum between the values after one-layer GCN-based graph convolution process of the existing beliefs and uncertainties $\hat{\bf B}^{(t-1)}$ and $\hat{\bf U}^{(t-1)}$ and their newly computed gates  ${\bf Z}_{\bf B}^{(t)}$ and ${\bf Z}_{\bf U}^{(t)}$. 

The reset gates ${\bf S}^{(t)}_{\bf B}$ and ${\bf S}^{(t)}_{\bf U}$ are designed to decide how much of the past information of beliefs and uncertainties 
should be reserved. Here, we apply another layer of GCN-based graph convolution process that extracts the information of beliefs and uncertainties to effectively capture their topological dependency information of the network structure, which is used as the input of the reset gate: 
\vspace{-1mm}
\begin{small}
\begin{eqnarray}
\left[{\bf S}^{(t)}_{\bf B}, {\bf S}^{(t)}_{\bf U}\right]= \sigma\Big(\Big[g_{\textbf{W}_\textbf{S}} \star [\hat{\bf B}^{(t-1)}, \textbf{X}^{(t)}],g_{\textbf{W}_\textbf{S}} \star [\hat{\bf U}^{(t-1)}, \textbf{X}^{(t)}]\Big]\Big)
\label{reset gates}
\end{eqnarray}
\end{small}
\vspace{-1mm}
where $\textbf{W}_\textbf{S}$ refers to the parameters of a graph convolution layer specific for reset gates. 



The beliefs and uncertainties ($\hat{\bf B}^{(t)}$ and $\hat{\bf U}^{(t)}$) at time $t$ are predicted based on the past information that are extracted by the reset gates (${\bf S}^{(t)}_{\bf B}$ and ${\bf S}^{(t)}_{\bf U}$) and the current information represented by the feature matrix $\textbf{X}^{(t)}$. In this step, we apply another layer of GCN-based graph convolutional process as a nonlinear filtering of the past information and the current information across the network topology:  
\begin{eqnarray}
\tilde{{\bf B}}^{(t)} &=& \tanh \Big(g_{\textbf{W}_{\tilde{{\bf B}}}} \star \Big[{\bf S}^{(t)}_{\textbf{B}}\odot \hat{\bf B}^{(t-1)}, \textbf{X}^{(t)}\Big]\Big),\\
\tilde{{\bf U}}^{(t)} &=& \tanh \Big(g_{\textbf{W}_{\tilde{{\bf U}}}} \star \Big[{\bf S}^{(t)}_{\textbf{U}}\odot \hat{\bf U}^{(t-1)}, \textbf{X}^{(t)}\Big]\Big), \label{last}
\end{eqnarray}
where we consider two different convolutions $g_{\textbf{W}_{\tilde{{\bf B}}}} \star$ and $g_{\textbf{W}_{\tilde{{\bf U}}}} \star$ for the candidate beliefs $\tilde{\bf B}^{(t)}$ and uncertainties $\tilde{\bf U}^{(t)}$ to model their own graph structural dependencies and $\textbf{W}_{\tilde{{\bf B}}}$ and $\textbf{W}_{\tilde{{\bf U}}}$ are their related parameters, respectively. 

Fig.~\ref{fig:vae-opinion-model} provides a graphical illustration of our proposed DL-based dynamic opinion model and the following provides the complete formulation of the model: 

\vspace{-1mm}
\begin{small}
\begin{eqnarray}
b^{(t)}_i &=& \hat{b}^{(t)}_i + e^{(t)}_i,\ e^{(t)}_i \sim \text{Lap}(\tau_b), i \in \mathbb{L}_t, \nonumber \\
u^{(t)}_i &=& \hat{u}^{(t)}_i + r^{(t)}_i,\ r^{(t)}_i \sim \text{Lap}(\tau_u), i \in \mathbb{L}_t, \nonumber \\
\hat{\bf B}^{(t)} &=&  {\bf Z}^{(t)}_{\bf B} \odot \hat{\bf B}^{(t-1)} + (1-{\bf Z}_{\bf B}^{(t)}) \odot \tilde{{\bf B}}^{(t)},\nonumber\\
\hat{\bf U}^{(t)} &=&  {\bf Z}^{(t)}_{\bf U} \odot \hat{\bf U}^{(t-1)} + (1-{\bf Z}^{(t)}_{\bf U}) \odot \tilde{{\bf U}}^{(t)}, \nonumber\\
\left[{\bf Z}_{\bf B}^{(t)}, {\bf Z}_{\bf U}^{(t)}\right] &=& \sigma\Big(\Big[g_{\textbf{W}_\textbf{Z}} \star [\hat{\bf B}^{(t-1)}, \textbf{X}^{(t)}],g_{\textbf{W}_\textbf{Z}} \star [\hat{\bf U}^{(t-1)}, \textbf{X}^{(t)}]\Big]\Big),\nonumber\\
\left[{\bf S}^{(t)}_{\bf B}, {\bf S}^{(t)}_{\bf U}\right]&=& \sigma\Big(\Big[g_{\textbf{W}_\textbf{S}} \star [\hat{\bf B}^{(t-1)}, \textbf{X}^{(t)}],g_{\textbf{W}_\textbf{S}} \star [\hat{\bf U}^{(t-1)}, \textbf{X}^{(t)}]\Big]\Big),\nonumber\\
\tilde{{\bf B}}^{(t)} &=& \tanh \Big(g_{\textbf{W}_{\hat{{\bf B}}}} \star [{\bf S}^{(t)}_{\textbf{B}}\odot \hat{\bf B}^{(t-1)}, \textbf{X}^{(t)}]]\Big),\nonumber\\
\tilde{{\bf U}}^{(t)} &=& \tanh \Big(g_{\textbf{W}_{\tilde{{\bf U}}}} \star [{\bf S}^{(t)}_{\textbf{U}}\odot \hat{\bf U}^{(t-1)}, \textbf{X}^{(t)}]]\Big).\nonumber
\end{eqnarray}
\end{small}
\vspace{-4mm}

\subsection{Inference Algorithm of Predicting Dynamic Uncertain Opinions}

Let $\Theta = \{\textbf{W}_\textbf{z}, \textbf{W}_\textbf{s}, \textbf{W}_{\tilde{{\bf B}}}, \textbf{W}_{\tilde{{\bf U}}} \}$ be the parameters of the proposed model. The log probability of the model based on the observed beliefs  and uncertainties $\{\textbf{B}_{\mathcal{L}_t}, \textbf{U}_{\mathcal{L}_t}\}_{t=1}^T$ can be formulated as follows:
\begin{gather}
\mathcal{L}(\Theta, \{\hat{\textbf{B}}_{\mathcal{L}_t}, \hat{\textbf{U}}_{\mathcal{L}_t}\}_{t=1}^T) \nonumber \\
= \log\prod_{t=1}^{T}\prod_{i \in \mathbb{L}_t} \left(  \text{Prob}(b_i^{(t)} | \hat{b}_i^{(t)}; \tau_b)  \text{Prob}(u_i^{(t)} | \hat{u}_i^{(t)}; \tau_u)\right)\nonumber \\
=  -\sum_{t=1}^{T}\sum_{i \in \mathbb{L}_t}\left( \frac{|b_i^{(t)} - \hat{b}_i^{(t)}|}{\tau_b} + \frac{|u_i^{(t)} - \hat{u}_i^{(t)}|}{\tau_u}\right) \label{loss},
\end{gather}
where the latent beliefs and uncertainties $\{\textbf{B}_{\mathcal{L}_t}, \textbf{U}_{\mathcal{L}_t}\}_{t=1}^T$ are calculated via Eqs.~\eqref{m1}-- \eqref{last}. 

Our inference algorithm for predicting the unknown dynamic opinions is designed to maximize the log probability function $\mathcal{L}(\Theta, \{\hat{\textbf{B}}_{\mathcal{L}_t}, \hat{\textbf{U}}_{\mathcal{L}_t}\}_{t=1}^T)$ over the parameters $\Theta$ and the latent beliefs and uncertainties $\{\hat{\textbf{B}}_{\mathcal{L}_t}, \hat{\textbf{U}}_{\mathcal{L}_t}\}_{t=1}^T$: 
\begin{gather}
\small
\max_{\Theta, \{\hat{\textbf{B}}_{\mathcal{L}_t}, \hat{\textbf{U}}_{\mathcal{L}_t}\}_{t=1}^T} \mathcal{L}(\Theta, \{\hat{\textbf{B}}_{\mathcal{L}_t}, \hat{\textbf{U}}_{\mathcal{L}_t}\}_{t=1}^T) \nonumber \\
= \min_{\Theta, \{\hat{\textbf{B}}_{\mathcal{L}_t}, \hat{\textbf{U}}_{\mathcal{L}_t}\}_{t=1}^T} \sum_{t=1}^{T}\sum_{i \in \mathbb{L}_t}\left( |b_i^{(t)} - \hat{b}_i^{(t)}| + \lambda \cdot |u_i^{(t)} - \hat{u}_i^{(t)}|\right),
\end{gather}
\normalsize
where $\lambda = \frac{\tau_b}{\tau_u}$ is a hyper-parameter to be tuned based on training data. Our proposed inference algorithm is designed using the framework of back propagation. As described in Algorithm~\ref{algorithm}, our proposed algorithm is an iterative alternating optimization procedure and in each iteration, a forward pass is designed to update the latent beliefs and uncertainties $\{\hat{\textbf{B}}_{\mathcal{L}_t}, \hat{\textbf{U}}_{\mathcal{L}_t}\}_{t=1}^T$ using the parameters $\Theta$ estimated in the previous iteration (Step 7). And then a backward pass is designed to update the parameters $\Theta$ by fixing the estimated latent beliefs and uncertainties $\{\hat{\textbf{B}}_{\mathcal{L}_t}, \hat{\textbf{U}}_{\mathcal{L}_t}\}_{t=1}^T$ (Steps 8 to 9). The key steps are described in Algorithm~\ref{algorithm}. 
After the latent beliefs and uncertainties $\{\hat{\textbf{B}}_{\mathcal{L}_t}, \hat{\textbf{U}}_{\mathcal{L}_t}\}_{t=1}^T$ are estimated, the unknown opinions $\{\bm{\omega}^{(t=1)}_{\mathbb{V}\setminus \mathbb{L}_{t=1}}, \cdots, \bm{\omega}^{(t=T)}_{\mathbb{V}\setminus \mathbb{L}_{t=T}}\}$ can be estimated by:
\begin{eqnarray}
\omega_i^{(t)} = \Big(\hat{b}_i^{t}, 1 - \hat{b}_i^{t} - \hat{u}_i^{(t)}, \hat{u}_i^{(t)} \Big), \ \ i \in \mathbb{L}_{\mathbb{V}, t=1, \cdots, T \setminus \mathbb{L}_{t}}\label{update-omega}
\end{eqnarray}
The conflicting opinions can also be identified by comparing the input beliefs and uncertainties $\{{\bf B}^{(t)}_{\mathbb{V} \setminus \mathbb{L}_t}, {\bf U}^{(t)}_{\mathbb{V} \setminus \mathbb{L}_t}\}_{t=1}^T$ and the estimated opinions  $\{\hat{\bf B}^{(t)}_{\mathbb{V} \setminus \mathbb{L}_t}, \hat{\bf U}^{(t)}_{\mathbb{V} \setminus \mathbb{L}_t}\}_{t=1}^T$. The nodes that have conflicting opinions at time $t$ can be identified as the set \[\left\{i \ | \ |b_i^{(t)} - \hat{b}_i^{(t)}| > 0 \text{ or } |u_i^{(t)} - \hat{u}_i^{(t)}| > 0, i \in \mathbb{L}_{t}\right\}.\] 
The forward pass takes $O(TN)$ and the backward pass takes $O(TM)$ due to the approximations based on Chevyshev polynomials in GCN~\cite{DBLP:journals/corr/KipfW16}, where $N$ and $M$ refer to the numbers of nodes and edges in the input network, respectively. The total algorithmic running time is hence $(O(L(TN + TM)))$, where $L$ is the number of iterations. As shown in our experiments, $L$ scales constant with respect to $M$ and accordingly our algorithm scales linearly with respect to the number of edges.

\begin{algorithm}[t]
\small{
\DontPrintSemicolon
\KwIn{$\mathbb{G} = (\mathbb{V}, \mathbb{E}, y)$ and $\{\bm{\omega}^{(t=1)}_{\mathbb{L}_{t=T}}, \cdots, \bm{\omega}^{(t=T)}_{\mathbb{L}_{t=T}}\}$}
\KwOut{$\{\bm{\omega}^{(t=1)}_{\mathbb{V}\setminus \mathbb{L}_{t=1}}, \cdots, \bm{\omega}^{(t=T)}_{\mathbb{V}\setminus \mathbb{L}_{t=T}}\}$,}
\SetKwBlock{Begin}{function}{end function}
{
  $\ell=1$;\;
  $P = 10$; (Set the hidden units number) \;  
  $\eta = 0.001$; (Set the learning rate) \;
  Initialize the parameters $\Theta$\;
  Initialize the states $\hat{\textbf{B}}^{(0)}=0, \hat{\textbf{U}}^{(0)}=0$;\;
 \Repeat{convergence }
 {
 
 Forward pass to compute $\hat{\textbf{B}}^{(t)}, \hat{\textbf{U}}^{(t)}$ via Eq.(\ref{m3} -- \ref{last}) for $t=1, \ldots, T$;\;
 Backward pass via the chain-rule the calculate the sub-gradient gradient: $g^{(\ell)} = \nabla_\Theta \mathcal{L}(\Theta)$  via Eq.~\eqref{loss}\;
 Update parameters using step size $\eta$ via 
  $\Theta^{(\ell+1)} = \Theta^{(\ell)} - \eta \cdot g^{(\ell)}$\; 

  $\ell = \ell + 1$;\;
 }
 Calculate $\{\bm{\omega}^{(t=1)}_{\mathbb{V}\setminus \mathbb{L}_{t=1}}, \cdots, \bm{\omega}^{(t=T)}_{\mathbb{V}\setminus \mathbb{L}_{t=T}}\}$ based on
 $\{\hat{\textbf{B}}^{(t)}, \hat{\textbf{U}}^{(t)}\}_{t=1}^T$ via Eq. (\ref{update-omega})
 

\Return{$\{\bm{\omega}^{(t=1)}_{\mathbb{V}\setminus \mathbb{L}_{t=1}}, \cdots, \bm{\omega}^{(t=T)}_{\mathbb{V}\setminus \mathbb{L}_{t=T}}\}$}
}
\caption{GCN-GRU based Opinion Prediction}\label{algorithm}
}
\end{algorithm}

\section{Results and Analysis} \label{sec:results_analysis}

\subsection{Experimental Settings}
\subsubsection{Semi-synthetic Epinions dataset} 
We use the \emph{Epinions} dataset~\cite{Epinions} representing a {\em who-trust-who} in an online social network. This is a directed network consisting of 47,676 users (i.e., vertices) and 467,468 relationships (i.e., edges). As there are no ground truth opinions available from the dataset, we use a benchmark simulation model~\cite{richardson2003trust} to generate synthetic opinions. The simulation model has the following main steps: 
\begin{itemize}
\item {\bf Initialization}: 10\% of the edges are randomly selected and set the trust of the edges to $1$'s meaning that trust is not symmetric (i.e., $i$ trusts $j$ does not necessarily mean $j$ trusts $i$) where $i$ and $j$ are users in a given directed network. 
\item {\bf Exploration}: 1,000 exploration steps are performed to update trust relationships via the following trust rule: 
\vspace{-2mm}
\begin{eqnarray}
\text{Trust} (a,b) = 1 \wedge \text{Trust} (b, c) = 1 \rightarrow \text{Trust} (b,c) = 1.\label{trust-rule}
\vspace{-2mm}
\end{eqnarray}
The exploration step is used to generate synthetic trust observations on the edges of the network. For each exploration step, we randomly select one edge, identify the rule instances associated with this edge, and generate one observation of the edge ($0$ or $1$) based on the probability of the rule instances, where $1$ refers to trust while $0$ refers to distrust. By repeating the exploration step 1,000 times, we generate a realization of trust relationships on the edges in the network, in which the observations of 1,000 randomly selected edges were generated while the other edges do not have any observations in this realization. We then conduct the 2nd realization based on the previous one by randomly selecting 5\% of the edges and swapping their most recent observations from $1$ to $0$ or from $0$ to $1$ that are considered as their new trust observations at the current realization. 1,000 exploration steps are conducted to generate observations to make them consistent with the trust rule. Following this procedure, we generate $T$ realizations. 
\item {\bf Performance evaluation}: After conducting the $T$ realizations, each edge then has up to $T$ trust observations and its opinion can be estimated based on its trust observations. We consider a set of candidate values of $T\in \{100, 200, \cdots, 500\}$ corresponding to different uncertainty ranges that will be explained below. In order to conduct performance evaluation for different sizes of a network, we randomly sample sub-networks with the number of nodes $N \in \{500, 1000, 5000, 10000\}$ from the original Epinions network, respectively. The {\em testing edges} are randomly selected from all the edges with the percentages (or test ratios) $\in \{10\%, 20\%, 30\%, 40\%, 50\%, 60\%\}$ and are predicted based on the known opinions of the other edges which are {\em training edges}. 
\end{itemize}


\subsubsection{Road traffic datasets} We collected live road  traffic data from June 1, 2013 to March 31, 2014 across two cities from INRIX~\cite{INRIX}, Washington D.C. and Philadelphia (PA), as summarized in Table~\ref{table:datasets}. The raw INRIX dataset collected live traffic speed information from trucks per five-minute interval. A road link has a live speed measurement at a specific time interval if it has at least one truck traversing this link at the time interval; otherwise, it will be a missing speed value. In addition, the reference speed information is given for each road link per hour. A reference speed is defined as the ``uncongested free flow speed'' for each road segment~\cite{referencespeed}. It is calculated based upon the 85-th percentile of the measured speed for all time periods over a few years, where the reference speed serves as a threshold separating two traffic states, \emph{congested} vs. \emph{uncongested}. The road traffic dataset for each of the two cities is based on the observations for 43 weeks in total. An hour is represented by a specific combination of hours of a day ($h\in \{6, 9, 12,\cdots, 21\}$), days of a week ($d\in \{1,2,3,4,5\}$), and weeks ($w\in \{1,2,\cdots, 43\}$): $(h,d,w)$. We only considered work days from Monday ($d=1$) to Friday ($d=5$) and hours from 6AM ($h=6$) to 10PM ($h=22$). 

\subsubsection{Social network datasets} We use the dataset of {\em Social Spammers} in the Evolving Multi-Relational Social Network Dataset~\cite{fakhraei2015collective}. This anonymized dataset was collected from the {\em Tagged.com} social network website. It contains 5.6 million (i.e., vertices) users and 858 million links between them. Each user has 4 features and is manually labeled as ``spammer" or ``not spammer." Each link represents an action between two users and includes a timestamp and a type. The timestamp includes information for 10 days. Per day, according to the timestamp, we collect onace per 4 hours, and use the algorithm in~\cite{fakhraei2015collective} to classify where a user is a ``spammer" or ``not spammer" to generate observations. If there is no action for one user in 4 hours, we define this user has no observation during the time interval.

\small
\begin{table}[!ht]
\vspace{-3mm}
\caption{Description of the four real-world datasets}
\vspace{-3mm}
\centering
\resizebox{\columnwidth}{!}{%
\begin{tabular}{|c||c|c|c|c|c|c|c|c| }\hline
Dataset name&\# nodes& \# edges &  \# weeks & \# snapshots (hours) in total \\ \hline 
\textbf{Epinions} & 477,468   & 8,477,468 &  - & 380 \\\hline
\textbf{Washington, D.C.} & 1,522   & 5,028 &  43& 3440\\\hline
\textbf{Philadelphia} (PA) & 607 & 1,772 & 43& 3440\\\hline
\textbf{Spammer} & 165,410   & 2,441,388 &  -& 60\\\hline
\end{tabular}}
\vspace{-3mm}\label{table:datasets}
\end{table}
\normalsize

\textbf{Preprocessing of the networks}: The congestion labels in DC and PA datasets relate to edges (i.e., road links), but not nodes (i.e., intersections). As our proposed approach is for node-level opinion inference, we converted the DC and PA road networks to new networks, in which each node represents a road link  while each edge indicates whether a node (a road link) is connected to another node (another road link) in the original network data. Note that the same preprocessing is also conducted for the Epinions dataset. 

\textbf{Ground truth time-dependent opinions (i.e., beliefs and uncertainties) of training and testing edges in each dataset}. For each road traffic dataset (for DC and PA), the opinion of a specific (training or testing) link $s$ at time (i.e., hour) $(h,d,w)$ is estimated based on the observations of the same hour in previous $K$ weeks $\{x_{s,h,d,w},x_{s,h,d,w-1}, \cdots, x_{s,h, d, w-K+1}\}$ as the evidence, where $x_{s,h,d,w}$ refers to the congestion observation ($0$ or $1$) of the link $s$ at hour $(h,d,w)$ where $K$ refers to the size of a predefined observation time window. Some of the observations were not available, as only a subset of the links were traversed by the delivery trucks. Denote by $K_s$ the number of observations within the $K$ weeks for the link $s$ and $0 \le K_s \le K$. The belief, disbelief, and uncertainty masses $b_s$, $d_s$, and $u_s$ of a specific link $s$ are estimated by: 
\begin{eqnarray} \label{est-uncertainty}
b_s &=& \left(\sum\nolimits_{t=0}^{K-1} x_{s,h,d,w-t} - W \cdot a\right)/(K_s+W) \nonumber \\
d_s &=& \left(K - \sum\nolimits_{t=0}^{K-1} x_{s,h,d,w-t} + W \cdot a\right)/(K_s+W)\nonumber \\ 
u_s &=& W/(K_s+W), 
\end{eqnarray}
where we set the non-informative prior weight (i.e., an amount of uncertain evidence with $W =2$) and the base rate (i.e., prior belief with $a=0.5$). As $T$ is the maximum number of possible observations a link can have within a time window of size $K$, it can be used to calculate a lower bound on the uncertainty of a link as $W/(K+W)$, and the upper bound will be $100\%$. For example, for $K = 38$, the range of uncertainties of the links is $[5\%, 100\%]$. 

For the Epinions dataset, $T$ realizations (time stamps) are made in total. For each link $i$, the ground truth opinion of this link at a specific time $t$ is calculated based the observations of the same link in previous $K$ time stamps ($\{t-K+1, t-K+2, \cdots, t\}$) as the evidence using equations similar to Eq.~\eqref{est-uncertainty}. With the $K$ time stamps, some of the observations may be missing and accordingly the calculated uncertainty may vary for different links and/or times.

\subsubsection{Injecting conflicting opinions} Since the dataset does not contain any conflicting opinions, we create synthetic conflicting opinions for the given input parameter $P$ (\% of opinions conflicting with their neighbors) as follows. We iterate the procedure $\text{r}(N\times T\times K\%)$ times to generate $r(N\times T\times K\%)$ conflicting opinions, where $r(\cdot)$ rounds a numerical number to an integer. Per iteration, we randomly pick a training link $i\in \mathbb{V}$ and time $t\in \{1, \cdots, T\}$, such that the opinion of this link at time $t$ is consistent with the opinions of its neighboring links. We then change its belief and disbelief such that they are most conflicting with the average belief and disbelief of its neighboring nodes at time $t$. For example, if the average belief and disbelief are $0.8$ and $0.1$, respectively, then the conflicting belief and disbelief will be set to $0.1$ and $0.8$, respectively. We consider $K = 0\%, 5\%, 10\%, 20\%$. 

\subsubsection{Parameter settings} The main parameters for all the datasets include $K$ as the size of time window and $TR$ as the test ratio (or \% of tested edges). We tested $K$ different window sizes with $K \in \{3, 6, 8, 11, 38\}$ corresponding to the the uncertainty ranges $[25\%, 100\%]$, $[20\%, 100\%]$, $[15\%, 100\%]$, and $[5\%, 100\%]$, respectively. Due to the space constraint, we only showed the results for $T=38$ and the uncertainty region $[15\%, 100\%]$. We found that similar trends of the results under different window sizes are observed. 
The values of $TR$ are set to $\{10\%, 20\%, 30\%, 40\%, 50\%, 60\%\}$. 

\subsubsection{Performance metrics} 
Based on Eq.~\eqref{est-uncertainty}, the uncertainty mass, $u_s$, for each training or testing edge is a known and constant value, $u$, after the window size $K$ is predefined, without the actual observations of this link. For this reason, our study using the road traffic datasets focuses on comparing our algorithm against its variants or counterparts with respect to the following metrics: {\em Belief Mean Absolute Error} (B-MAE), {\em Uncertainty Mean Absolute Error} (U-MAE), and computation time (in sec.). 

The B-MAE and U-MAE are calculated by: 
\begin{eqnarray}
\text{B-MAE}({\bm \omega}_{\mathbb{V} \setminus \mathbb{L}}) = \frac{1}{N}\sum\nolimits_{i\in \mathbb{V} \setminus \mathbb{L}}  \left|b_i - b_i^\star\right|  \\
\text{U-MAE}({\bm \omega}_{\mathbb{V} \setminus \mathbb{L}}) = \frac{1}{N}\sum\nolimits_{i \in \mathbb{V} \setminus \mathbb{L}}  \left|u_i - u_i^\star\right|  
\end{eqnarray} 
\noindent where $\omega_i = (b_i, d_i, u_i, a)$ and $\omega_j^\star = (b_i^\star, d_i^\star, u_i^\star, a)$ refer to the predicted and true opinions of a target variable $y_i$ associated with node $i$, respectively. To accurately estimate the running complexity, we use the computation time to represent the efficiency of algorithms evaluated in this work. But to compare the asymptotic algorithmic complexity of comparing algorithms, we also summarize the {\em Big-O} of considered algorithms in Table~\ref{table:big-o}.

\subsubsection{Comparison methods} We notate our proposed DL-based dynamic opinion model as {\em GCN-GRU-opinion}. We compared our proposed methods with the comparable two counterpart methods: {\em SL}~\cite{josang2016subjective}, {\em CSL}~\cite{chen2017collective}, {\em GRU-opinion}~\cite{DBLP:journals/corr/ChoMGBSB14} and {\em GCN-semi} for semi-supervised node classification~\cite{DBLP:journals/corr/KipfW16}. Note that {\bf CSL} is not directly comparable to our proposed methods because CSL was designed for the scenario where all the node-level opinions in a network have the same uncertainties but different beliefs (or disbeliefs). However, in this work, we consider varying uncertainties across nodes. We employed the following procedure to predict the missing values for the training edges, such that CSL can be considered: Per road link $i$, we first estimated its opinion based on its available observations within the size of the current time window $T$, and then used its equivalent Beta PDF to sample binary observations for its missing observations within the time window. After this procedure, each training edge has the number of observations $T$ and hence the same uncertainty values. Another baseline method, {\bf GRU-opinion} was designed for predicting dynamic opinions where we input the observed node opinion to predict unobserved node opinion. {\bf GCN-Semi} was designed for semi-supervised node classification, but not directly for opinion inference. We made the following modifications to adapt GCN-Semi for opinion inference: For each time interval within a time window $T$, we applied GCN-Semi to predict the congestion labels of the testing edges and the missing labels of the training edges simultaneously. Then, for each testing edge, we obtained the corresponding $T$ observations within $T$ time window. Then we used these observations to directly estimate its opinion. Following this strategy, as all the testing links have the same number of observations ($T$), their predicted uncertainties will be identical. For the running complexity of algorithms considered in this work, we summarize a respective {\em Big-O} in Table~\ref{table:big-o}.

\begin{table}[!ht]
\vspace{-2.5mm}
\caption{Summary of algorithmic complexity of comparing algorithms in {\em Big-O}: For a given network, $L$ is the number of iterations, $T$ is the number of time intervals, $N$ is the number of nodes, $M$ is the number of edges, and $R$ is the total number of shortest paths calculated in SL that scales exponentially with respect to $N$.}
\vspace{-2mm}
\centering
\resizebox{\columnwidth}{!}{%
\begin{tabular}{|c||c|c|c|c|}\hline
GCN-GRU-opinion & GRU-opinion & GCN-Semi & CSL & SL \\ \hline 
$O(LT(N+M))$ & $O(LT(N))$ & $O(LT(N+M))$ & $O(LT(N+M))$ & $O(TNR)$ \\ \hline
\end{tabular}}
\vspace{-8mm}\label{table:big-o}
\end{table}
\subsubsection{Parameter tuning} SL only has one hyperparameter that is the maximum length of its independent paths. We set this to $18$ as the results of SL are obaserved almost the same for the maximum lengths equal to or greater than $18$. CSL does not have hyperparameters for tuning. Our proposed methods, GCN-GRU-opinion, GRU-opinion, GRU have three hyper parameters: $\lambda$ for tuning a trade-off, $\eta$ for a learning rate, and $P$ for the hidden units. We set $\lambda = 1, \eta = 0.01$, $P=16$, and $dropout = 0.2$ for all the experiments. All these hyperparameters are estimated based on the observations of the training edges. 

\begin{figure*}[!htb]
  \centering
  \subfigure[B-MAE, Epinion Dataset]{
    \includegraphics[width=0.23\textwidth, height=0.14\textwidth]{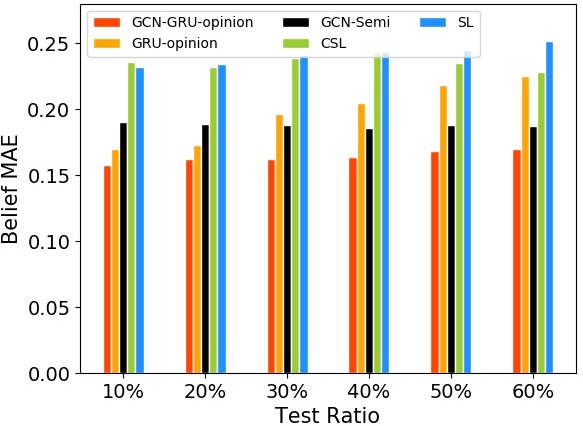}}
  \subfigure[U-MAE]{
    \includegraphics[width=0.23\textwidth, height=0.14\textwidth]{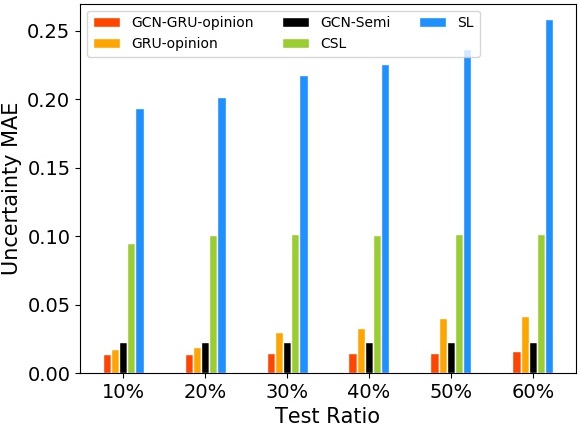}}
  \subfigure[B-MAE]{
    \includegraphics[width=0.23\textwidth, height=0.14\textwidth]{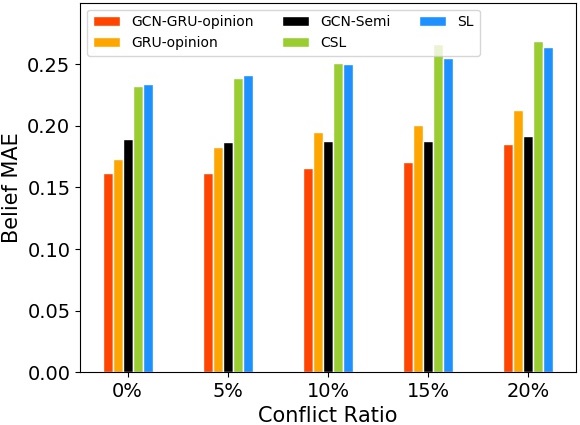}}
  \subfigure[U-MAE]{
    \includegraphics[width=0.23\textwidth, height=0.14\textwidth]{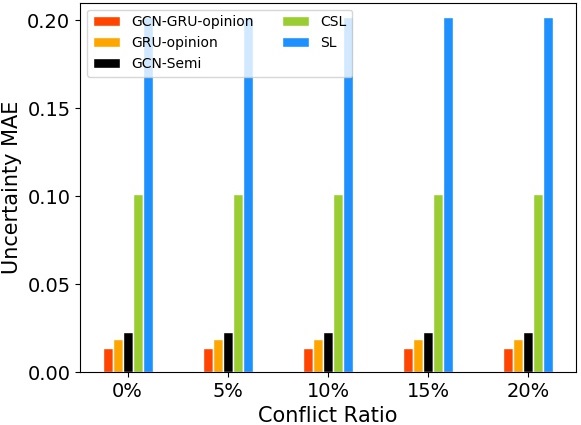}}   
    \vspace{-2mm}
  \caption{Comparison between the proposed algorithm (GCN-GRU-opinion) and its counterparts: belief and uncertainty MAEs vs. test ratios and conflict ratios based on the Epinion dataset.}
  \label{fig:epinion}
  \vspace{-3mm}
\end{figure*}

\subsection{Experimental Results based on Semi-Synthetic Datasets}
Fig.~\ref{fig:epinion} shows the comparative analysis of our proposed GCN-GRU-opinion and the four counterpart methods using the semi-synthetic Epinions dataset, with respect to Belief-MAE and Uncertainty-MAE. Fig.~\ref{fig:epinion} (a) and (b) demonstrates the effect of test ratio on both MAE metrics of all compared schemes. Obviously, GCN-GRU-opinion outperforms among all in both MAE metrics, except that they perform comparably to GRU-opinion for the setting ($TR = 10\%$) for Uncertainty-MAE. Both Belief-MAE and Uncertainty-MAE do not show clear sensitivity on different test ratios except MAE metrics for SL. The reason of the little sensitivity over the ranges of varying test ratios is as follows. For CSL and GCN-semi, for given $T$, the uncertainty of each edge (either testing or training) is assumed to be a constant (i.e., $(W/(W+T)$), which leads little sensitivity of uncertainty across different test ratios. For GCN-GRU-opinion, the little sensitivity shows high resilience even with a small testing ratio because it can predict better due to the nature of two layers of graph convolutional process based on GCN that can allows the maximization of prediction accuracy. 

Fig.~\ref{fig:epinion} (c) and (d) shows the effect of a conflict noise on Belief-MAE and Uncertainty-MAE under all comparing schemes. It is clear that GCN-GRU-opinion outperform their counterparts. There is an interesting pattern that when a conflict noise increases, GCN-GRU-opinion shows decreasing Belief-MAE while GCN-Semi gives a stable performance. It implies that the GCN-GRU-opinion and GCN-Semi can properly deal with the noise by leveraging the two layers of graph convolutional process based on GCN.

\begin{figure*}[!htb]
  \centering
  \subfigure[B-MAE, PA Dataset]{
    \includegraphics[width=0.23\textwidth, height=0.14\textwidth]{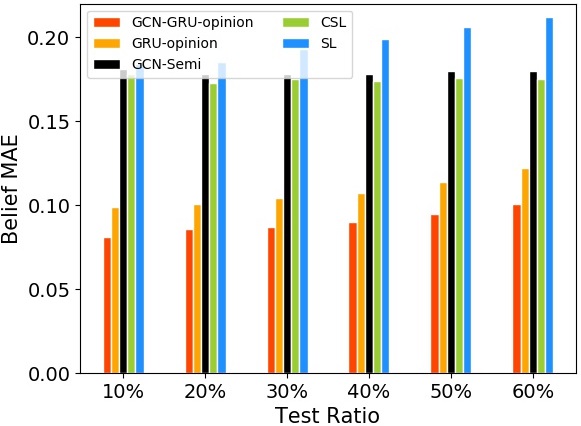}}
  \subfigure[U-MAE, PA Dataset]{
    \includegraphics[width=0.23\textwidth, height=0.14\textwidth]{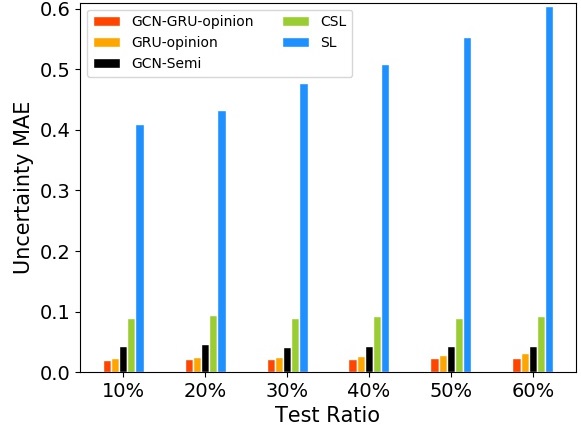}}
  \subfigure[B-MAE, DC Dataset]{
    \includegraphics[width=0.23\textwidth, height=0.14\textwidth]{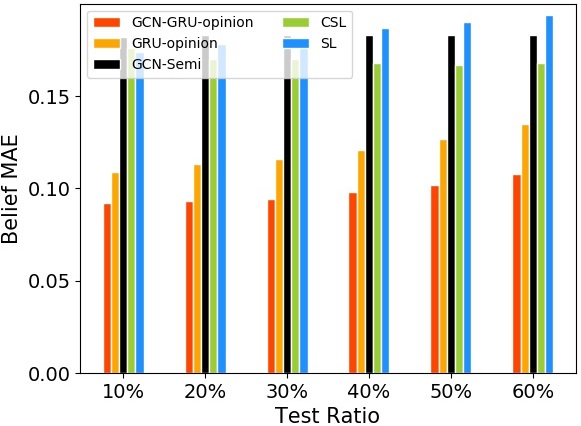}}
  \subfigure[U-MAE, DC Dataset]{
    \includegraphics[width=0.23\textwidth, height=0.14\textwidth]{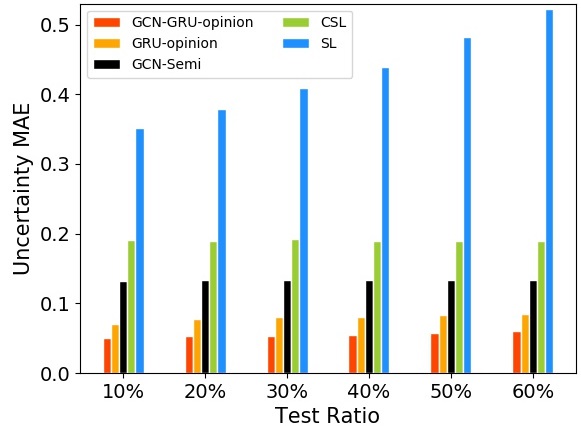}}   
    \vspace{-2mm}
  \caption{Comparison between our proposed algorithm (GCN-GRU-opinion) and its counterparts: belief and uncertainty MAEs vs. test ratios based on traffic road datasets.}
  \label{fig_accu}
  \vspace{-2mm}
\end{figure*}

\begin{figure*}[!htb]
  \centering
  \subfigure[B-MAE, PA Dataset]{
    \includegraphics[width=0.23\textwidth, height=0.14\textwidth]{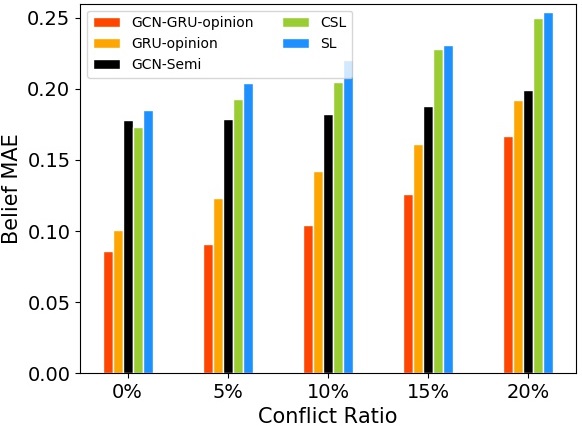}}
  \subfigure[U-MAE, PA Dataset]{
    \includegraphics[width=0.23\textwidth, height=0.14\textwidth]{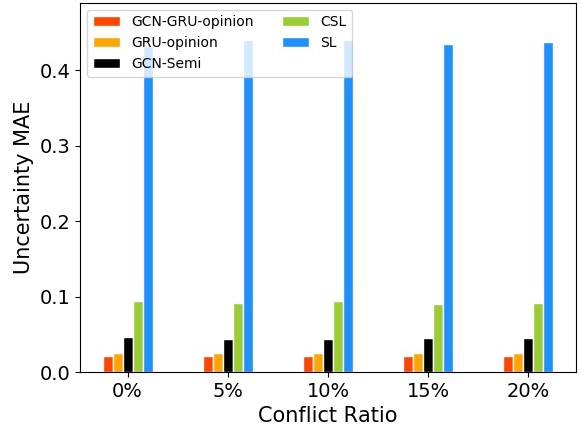}}
  \subfigure[B-MAE, DC Dataset]{
    \includegraphics[width=0.23\textwidth, height=0.14\textwidth]{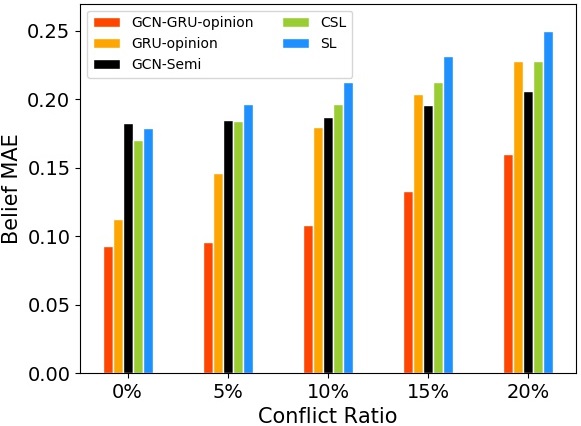}}
  \subfigure[U-MAE, DC Dataset]{
    \includegraphics[width=0.23\textwidth, height=0.14\textwidth]{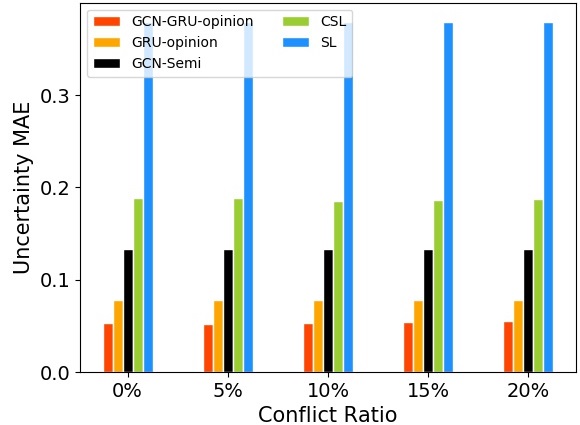}}
    \vspace{-2mm}
  \caption{Comparison between our proposed algorithms (GCN-GRU-opinion) and its counterparts: belief and uncertainty MAEs vs. conflict ratios based on traffic road datasets.}
  \label{fig_conflic_trffic}
  \vspace{-2mm}
\end{figure*}

\begin{figure}[!htb]
  \begin{center}
  \vspace{-2mm}   
  \includegraphics[width=0.23\textwidth, height=0.14\textwidth]{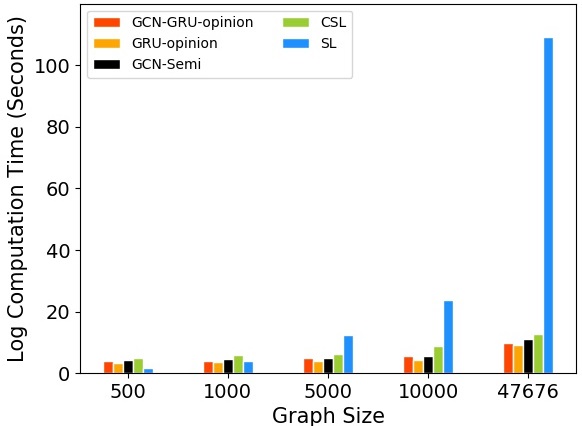}
  \end{center}
  \vspace{-5mm}
  \caption{Computation time based on the Epinion dataset with respect to increasing network size.}
  \vspace{-5mm}
\label{fig:epinion-runningtime}
\end{figure}

Fig.~\ref{fig:epinion-runningtime} shows the logarithmic computation times as the number of nodes increases. Except SL whose computation time increases exponentially when the network size increases, other schemes scale nearly linearly with respect to the network size. GCN-GRU-opinion and GRU-opinion are the most efficient methods among all the methods. More discussions about the computation times of these methods are presented in the below section for the real-world datasets. 

\subsection{Experimental Results based on Traffic Road Datasets}

Fig.~\ref{fig_accu} compares the performances of our proposed GCN-GRU-based algorithm with those of the four counterpart methods (i.e., CSL, SL, GRU-opinion, and GCN-Semi) with respect to Belief-MAE and Uncertainty-MAE based on two road traffic datasets (PA and DC). The results indicate that GCN-GRU-opinion performs the best among all in both prediction of beliefs and uncertainties. GRU-opinion performs the second best, but it is much more sensitive across different test ratios than GCN-GRU-opinion, showing that both Belief-MAE and uncertainty-MAE in GRU-opinion significantly increase as the test ratio increases. As the test ratio increases, the performance of the GCN-GRU-opinion is more pronounced showing the higher gap compared to the GRU-opinion.

Fig.~\ref{fig_conflic_trffic} shows the performance of comparing schemes with respect to different conflict ratio. Obviously, the GCN-GRU-opinion outperforms among all in predicting both beliefs and uncertainties. Unlike the little sensitivity of increasing the test ratio in comparing schemes, the GCN-GRU-opinion shows higher sensitivities across different conflict ratios with respect to the Belief-MAE.

\subsection{Experimental Results based on Social Network Dataset}

\begin{figure*}[!htb]
  \centering
  \subfigure[B-MAE]{
    \includegraphics[width=0.23\textwidth, height=0.14\textwidth]{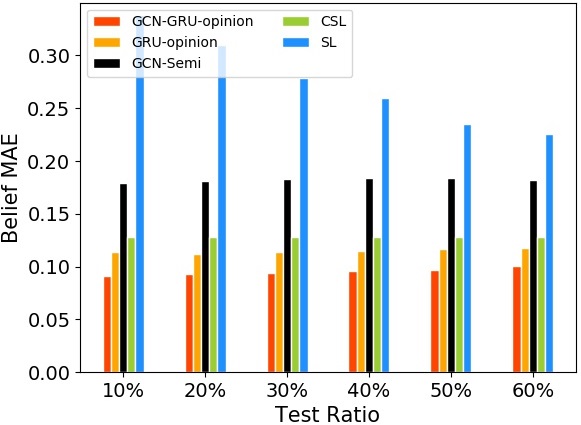}}
  \subfigure[U-MAE]{
    \includegraphics[width=0.23\textwidth, height=0.14\textwidth]{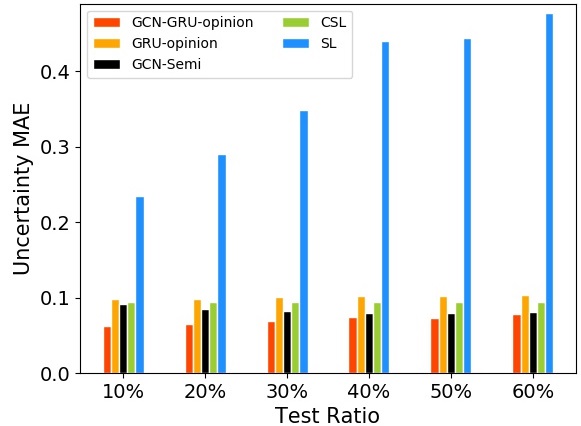}}
  \subfigure[B-MAE]{
    \includegraphics[width=0.23\textwidth, height=0.14\textwidth]{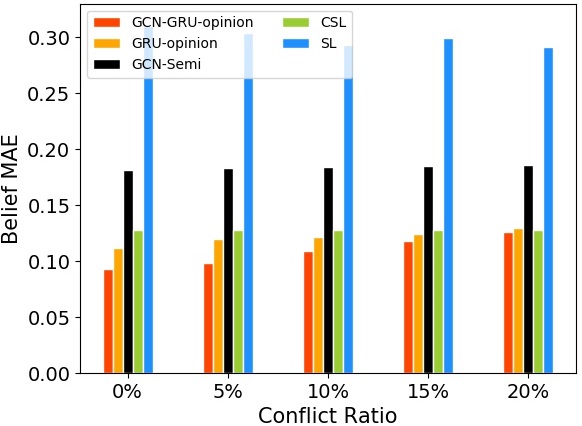}}
  \subfigure[U-MAE]{
    \includegraphics[width=0.23\textwidth, height=0.14\textwidth]{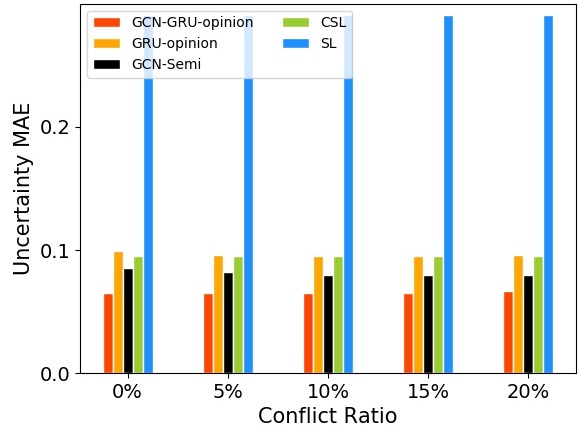}}
    \vspace{-2mm}
  \caption{Comparison of our proposed methods (GCN-GRU-opinion) and counterpart methods: Belief and Uncertainty MAEs vs.Test Ratios adn Conflict Ratios on Spammer Dataset}
  \label{fig_spam}
  \vspace{-5mm}
\end{figure*}
Fig.~\ref{fig_spam} shows the comparative analysis of our proposed GCN-GRU-opinion and the four counterpart methods {in terms of} Belief-MAE and Uncertainty-MAE under the social network {Spammer} dataset. Fig.~\ref{fig_spam} (a) and (b) demonstrates the effect of test ratio on both MAE metrics of all compared schemes. Obviously, GCN-GRU-opinion outperform among {in} all both the MAE metrics, except that they perform comparable to CSL for some settings (TR = 40\%, 50\%, 60\%) for Uncertainty-MAE. 

Fig.~\ref{fig_spam} (c) and (d) shows the effect of a conflict noise on Belief-MAE and Uncertainty-MAE {under all comparing schemes}. It is clear that GCN-GRU-opinion outperform their counterparts on both the metrics. There is an interesting pattern that when a conflict noise increases, Belief-MAE {decreases} for GCN-GRU-opinion and GCN-Semi shows a stable performance. It implies that the GCN-GRU-opinion and GCN-Semi can defend the noise.

\begin{figure}[!htb]
  \begin{center}
  \vspace{-2mm}   
  \includegraphics[width=0.29\textwidth, height=0.17\textwidth]{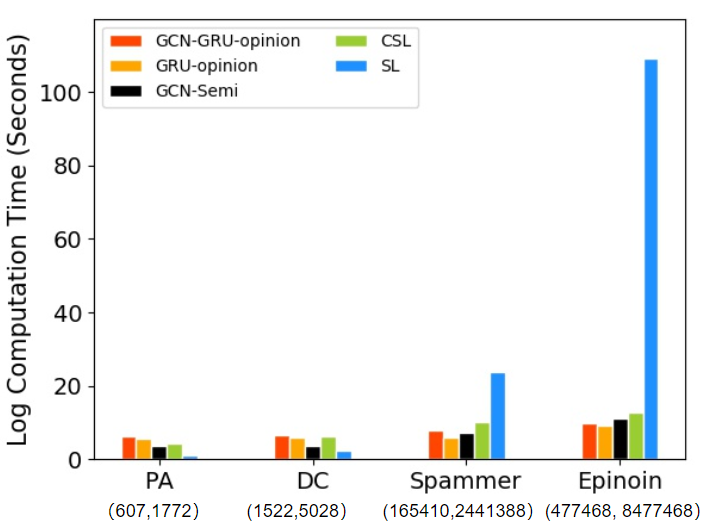}
  \end{center}
  \vspace{-5mm}
  \caption{Comparison of computation time on different dataset }
  \vspace{-8mm}
\label{fig:runningtime}
\end{figure}

Fig.~\ref{fig:runningtime} shows the average snapshot-level logarithmic computation times across different datasets. When the network size increases, the running complexity of SL exponentially increases while those of the rest of comparing algorithms linearly increase. In particular, GRU-opinion and GCN-Semi show the smallest computation time. GCN-GRU-opinion shows a lower computation time than CSL and SL, but a higher computation time than GRU-opinion and GCN-Semi due the complexity caused by combining the GCN-based model with GRU-based model. However, GCN-GRU-opinion still scales almost linearly in proportion to a network size. CSL and SL introduce the longest computation time, because the model needs to run on the snapshot of the network for each time slot while the other methods only run once for the whole time series data. 

\section{Conclusion and Future Work} \label{sec:conclusion-future-work}

In this work, we propose a novel DL-based dynamic opinion approach based on GCN and GRU techniques to address the key challenges of scalability and handling dynamic opinions in network data. From the simulation experiments conducted in this work, we can summarize the {\bf key findings} as follows:
\begin{itemize}
\item Overall our proposed GCN-GRU-opinion algorithm outperforms among all comparing counterparts in both Belief-MAE and Uncertainty-MAE. In particular, our GCN-GRU-opinion shows less sensitivity over a wide range of test ratios, implying higher resilience.
\item The performance order in Belief-MAE follows: GCN-GRU-opinion $>$ GRU-opinion $>$ GCN-Semi $>$ CSL $>$ SL. The performance order in Uncertainty-MAE follows: GCN-GRU-opinion $>$ GRU-opinion $>$ GCN-Semi $\approx$ CSL $>$ SL.
\item The higher performance of GRU-based methods is because they are capable of modeling dynamic structural dependencies among node-level beliefs and uncertainties.
\item The higher performance of GCN-GRU-opinion over GCN-GRU is because GCN-GRU-opinion integrates an GRU-based opinion model to model the time series relational dependencies between beliefs and uncertainties based on dynamic structural dependencies.
\item 
GCN-GRU-opinion scales in a close-to-linear complexity in proportion to the network size, proving high scalability under large-scale network data.
\end{itemize}

{\small

\begin{thebibliography}{21}
\providecommand{\natexlab}[1]{#1}
\providecommand{\url}[1]{#1}
\csname url@samestyle\endcsname
\providecommand{\newblock}{\relax}
\providecommand{\bibinfo}[2]{#2}
\providecommand{\BIBentrySTDinterwordspacing}{\spaceskip=0pt\relax}
\providecommand{\BIBentryALTinterwordstretchfactor}{4}
\providecommand{\BIBentryALTinterwordspacing}{\spaceskip=\fontdimen2\font plus
\BIBentryALTinterwordstretchfactor\fontdimen3\font minus
  \fontdimen4\font\relax}
\providecommand{\BIBforeignlanguage}[2]{{%
\expandafter\ifx\csname l@#1\endcsname\relax
\typeout{** WARNING: IEEEtranSN.bst: No hyphenation pattern has been}%
\typeout{** loaded for the language `#1'. Using the pattern for}%
\typeout{** the default language instead.}%
\else
\language=\csname l@#1\endcsname
\fi
#2}}
\providecommand{\BIBdecl}{\relax}
\BIBdecl

\bibitem[Epi()]{Epinions}
``Epinions,'' \url{http://www.trustlet.org/downloaded_epinions.html}.

\bibitem[INR()]{INRIX}
``Inrix,'' \url{http://inrix.com/publicsector.asp}.

\bibitem[ref()]{referencespeed}
``Reference speed for congestion evaluation,''
  \url{http://www.inrix.com/scorecard/methodology.asp/}.

\bibitem[Bach et~al.(2015)Bach, Broecheler, Huang, and Getoor]{bach2015hinge}
S.~H. Bach, M.~Broecheler, B.~Huang, and L.~Getoor, ``Hinge-loss markov random
  fields and probabilistic soft logic,'' \emph{arXiv preprint
  arXiv:1505.04406}, 2015.

\bibitem[Bruna et~al.(2013)Bruna, Zaremba, Szlam, and LeCun]{bruna2013spectral}
J.~Bruna, W.~Zaremba, A.~Szlam, and Y.~LeCun, ``Spectral networks and locally
  connected networks on graphs,'' \emph{arXiv preprint arXiv:1312.6203}, 2013.

\bibitem[Chen et~al.(2017)Chen, Wang, and Cho]{chen2017collective}
F.~Chen, C.~Wang, and J.-H. Cho, ``Collective subjective logic: Scalable
  uncertainty-based opinion inference,'' in \emph{IEEE BigData}, 2017, pp. 7--16.

\bibitem[Cho et~al.(2017)Cho, Cook, Rager, O'Donovan, and Adali]{Cho17}
J.~Cho, T.~Cook, S.~Rager, J.~O'Donovan, and S.~Adali, ``Modeling and analysis
  of uncertainty-based false information propagation in social networks,'' in
  \emph{2017 IEEE GLOBECOM}, Singapore, Dec. 2017, pp. 1--7.

\bibitem[Cho et~al.(2014)Cho, van Merrienboer, G{\"{u}}l{\c{c}}ehre, Bougares,
  Schwenk, and Bengio]{DBLP:journals/corr/ChoMGBSB14}
\BIBentryALTinterwordspacing
K.~Cho, B.~van Merrienboer, {\c{C}}.~G{\"{u}}l{\c{c}}ehre, F.~Bougares,
  H.~Schwenk, and Y.~Bengio, ``Learning phrase representations using {RNN}
  encoder-decoder for statistical machine translation,'' \emph{CoRR}, vol.
  abs/1406.1078, 2014. [Online]. Available:
  \url{http://arxiv.org/abs/1406.1078}
\BIBentrySTDinterwordspacing

\bibitem[Fakhraei et~al.(2015)Fakhraei, Foulds, Shashanka, and
  Getoor]{fakhraei2015collective}
S.~Fakhraei, J.~Foulds, M.~Shashanka, and L.~Getoor, ``Collective spammer
  detection in evolving multi-relational social networks,''in
  \emph{KDD}.\hskip 1em plus 0.5em minus 0.4em\relax ACM, 2015, pp. 1769--1778.

\bibitem[Frasconi et~al.(1998)Frasconi, Gori, and
  Sperduti]{frasconi1998general}
P.~Frasconi, M.~Gori, and A.~Sperduti, ``A general framework for adaptive
  processing of data structures,'' \emph{IEEE transactions on Neural Networks},
  vol.~9, no.~5, pp. 768--786, 1998.

\bibitem[Gori et~al.(2005)Gori, Monfardini, and Scarselli]{gori2005new}
M.~Gori, G.~Monfardini, and F.~Scarselli, ``A new model for learning in graph
  domains,'' in \emph{Neural Networks, 2005. IJCNN'05. Proceedings. 2005 IEEE
  International Joint Conference on}, vol.~2.\hskip 1em plus 0.5em minus
  0.4em\relax IEEE, 2005, pp. 729--734.

\bibitem[Hamilton et~al.(2017)Hamilton, Ying, and
  Leskovec]{hamilton2017inductive}
W.~Hamilton, Z.~Ying, and J.~Leskovec, ``Inductive representation learning on
  large graphs,'' in \emph{NIPS},
  2017, pp. 1025--1035.

\bibitem[Hammond et~al.(2011)Hammond, Vandergheynst, and Gribonval]{Hammond11}
D.~K. Hammond, P.~Vandergheynst, and R.~Gribonval, ``Wavelets on graphs via
  spectral graph theory,'' \emph{ACHA},
  vol.~30, no.~2, pp. 129--150, 2011.

\bibitem[Henaff et~al.(2015)Henaff, Bruna, and LeCun]{henaff2015deep}
M.~Henaff, J.~Bruna, and Y.~LeCun, ``Deep convolutional networks on
  graph-structured data,'' \emph{arXiv preprint arXiv:1506.05163}, 2015.

\bibitem[Ivanovska et~al.(2015)Ivanovska, J{\o}sang, Kaplan, and
  Sambo]{ivanovska2015subjective}
M.~Ivanovska, A.~J{\o}sang, L.~Kaplan, and F.~Sambo, ``Subjective networks:
  Perspectives and challenges,'' in \emph{GSKPR}.\hskip 1em plus 0.5em
  minus 0.4em\relax Springer, 2015, pp. 107--124.

\bibitem[J{\o}sang(2001)]{josang2001logic}
A.~J{\o}sang, ``A logic for uncertain probabilities,'' \emph{International
  Journal of Uncertainty, Fuzziness and Knowledge-Based Systems}, vol.~9,
  no.~03, pp. 279--311, 2001.

\bibitem[J{\o}sang(2016)]{josang2016subjective}
------, \emph{Subjective Logic: A Formalism for Reasoning Under
  Uncertainty}.\hskip 1em plus 0.5em minus 0.4em\relax Springer, 2016.

\bibitem[Kipf and Welling(2016)]{DBLP:journals/corr/KipfW16}
\BIBentryALTinterwordspacing
T.~N. Kipf and M.~Welling, ``Semi-supervised classification with graph
  convolutional networks,'' \emph{CoRR}, vol. abs/1609.02907, 2016. [Online].
  Available: \url{http://arxiv.org/abs/1609.02907}
\BIBentrySTDinterwordspacing

\bibitem[Pearl(2014)]{pearl2014probabilistic}
J.~Pearl, \emph{Probabilistic reasoning in intelligent systems: networks of
  plausible inference}.\hskip 1em plus 0.5em minus 0.4em\relax Morgan Kaufmann,
  2014.

\bibitem[Rasmussen(2004)]{rasmussen2004gaussian}
C.~E. Rasmussen, ``Gaussian processes in machine learning,'' in \emph{Advanced
  lectures on machine learning}.\hskip 1em plus 0.5em minus 0.4em\relax
  Springer, 2004, pp. 63--71.

\bibitem[Richardson et~al.(2003)Richardson, Agrawal, and
  Domingos]{richardson2003trust}
M.~Richardson, R.~Agrawal, and P.~Domingos, ``Trust management for the semantic
  web,'' in \emph{International semantic Web conference}.\hskip 1em plus 0.5em
  minus 0.4em\relax Springer, 2003, pp. 351--368.

\end{thebibliography}

}

\end{document}